\documentclass[conference]{IEEEtran}
\IEEEoverridecommandlockouts
\usepackage{cite}
\usepackage{amsmath,amssymb,amsfonts}
\usepackage{algorithmic}
\usepackage{graphicx}
\usepackage{textcomp}
\usepackage[table]{xcolor}

\usepackage[hidelinks]{hyperref}
\def\BibTeX{{\rm B\kern-.05em{\sc i\kern-.025em b}\kern-.08em
    T\kern-.1667em\lower.7ex\hbox{E}\kern-.125emX}}

\usepackage{multirow}

\usepackage[columnwise, switch]{lineno}

\begin{document}

\title{Advancing Perception in Artificial Intelligence through Principles of Cognitive Science}


\author{
Palaash~Agrawal,  ~Cheston~Tan ~and  ~Heena~Rathore
  \thanks{P. Agrawal and C. Tan are with the Center for Frontier AI Research, Agency for Science, Technology and Research (A*STAR), Singapore (e-mail: agrawal\_palaash@cfar.a-star.edu.sg; cheston\_tan@cfar.a-star.edu.sg).}
  \thanks{H. Rathore is with the Dept of Computer Science, Texas State University, Texas, USA (e-mail: heena.rathore@txstate.edu).}
  }

\maketitle

\begin{abstract}
Although artificial intelligence (AI) has achieved many feats at a rapid pace, there still exist open problems and fundamental shortcomings related to performance and resource efficiency. Since AI researchers benchmark a significant proportion of performance standards through human intelligence, cognitive sciences-inspired AI is a promising domain of research. Studying cognitive science can provide a fresh perspective to building fundamental blocks in AI research, which can lead to improved performance and efficiency. In this review paper, we focus on the cognitive functions of perception, which is the  process of taking signals from one's surroundings as input, and processing them to understand the environment. Particularly, we study and compare its various processes through the lens of both cognitive sciences and AI. Through this study, we review all current major theories from various sub-disciplines of cognitive science (specifically neuroscience, psychology and linguistics), and draw parallels with theories and techniques from current practices in AI.  We, hence, present a detailed collection of methods in AI for researchers to build AI systems inspired by cognitive science. Further, through the process of reviewing the state of cognitive inspired AI, we point out many gaps in the current state of AI (with respect to the performance of the human brain), and hence present potential directions for researchers to develop better perception systems in AI. 

\end{abstract}


\section{\textbf{Introduction}}

Artificial intelligence (AI) research has seen significant leaps in the last decade~\cite{glenskistate}. 
However, there are still many limitations to the various approaches used in AI, such as efficient storing and retrieval of vast amounts of information from limited computational resources \cite{liang2023brain}, the dependence of present AI methods on large amounts of data and processing power \cite{sarfraz2023towards}, and the ability to perform complex, layered and abstract reasoning. In order to overcome these limitations, it is worthwhile to revisit the definition and characteristics of intelligence. Understanding the nature of intelligence itself is a crucial building block to the realization of artificial general intelligence \cite{wang2021artificial, zhao2023brain}. 

\begin{figure}
    \centering
    \includegraphics[width = \columnwidth]{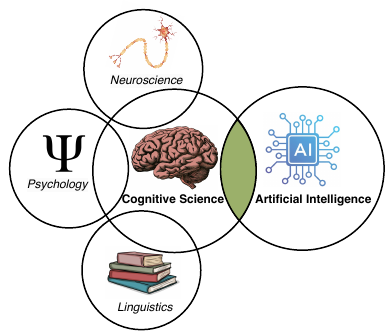}
    \caption{Cognitive-AI is the intersection of cognitive sciences and artificial intelligence. Cognitive sciences itself is a multidisciplinary field, and the intersection of all these fields with AI (the green region) is of great importance to understand intelligence. We study the state of cognitive AI from a fresh perspective in this paper. }
    \label{fig:CogAI representation}
\end{figure}

Among the various definitions of intelligence, one of the most widely accepted premises is that intelligence is cognitive in nature, i.e., it is defined through the benchmarking of human performance and behavior \cite{boltuc2020consciousness}. To this end, cognitive science inspired modeling in AI is seen as a  promising approach to advance the state of AI 
\cite{luo2022special}. Cognitive modeling focuses on deriving functional and structural mechanisms from the human brain along with its behavior and applying them to AI models~\cite{sangaiah2022cognitive, makkar2022securecps} (see Figure \ref{fig:CogAI representation}). By understanding cognitive modeling principles, one may understand the nature of intelligence in various forms such as logic, functionality, ability to solve complex problems, social behavior and ethics \cite{langley2022theory}.  Cognitive theory can help intelligent agents represent knowledge and reasoning logic more comprehensively~\cite{funge1999cognitive, kelley2018cognitive}. 
Many works in AI research have in fact shown significant improvement in performance by modeling based on cognitive principles. For example, a recent work showed that a modular model performed well on various spatial navigation tasks \cite{gervet2023navigating}.

\subsection{Perception and the state of AI}
The brain consists of various cognitive functional modules, such as perception, attention and memory. While all these have their own significance, perception 
is the gateway to all information collected by both the brain as well as AI models. It is, hence, highly fundamental to intelligence, compared to other cognitive functions. 
Perception is the process by which an agent (biological or artificial) gathers raw sensory information from its environment in various forms (modalities), and processes them in order to derive meaningful information (features) \cite{efron1969perception}. 
In the brain, all other cognitive functions build on top of sensory signals that are gathered and processed through the various perceptual pathways.
Further, various learning processes in both biological and AI systems heavily rely on the nature and quality of the information signals collected from their respective surroundings, thus making perception the bottleneck to intelligence. Hence, as AI continues to advance, improving perceptual processes is a critical step towards creating more capable and intelligent systems \cite{tian2017towards}. Incorporating cognitive science principles into models in AI can lead to AI systems that are not only more capable but also more aligned with human cognition and behavior \cite{yudkowsky2016ai}. 

The state of perception in AI is brittle and full of various kinds of open problems, which limit the scope of application to new problems. From a high level overview, visual perception (in the field of computer vision) still suffers from problems such as susceptibility to noisy and adversarial input data \cite{long2022survey}, long-tail recognition \cite{feldman2020neural}, the inability of models to learn from limited data \cite{parnami2022learning}, and generalizing from two-dimensional visual input to the semantics of three-dimensional visualizations \cite{yuniarti2019review}. Apart from these fundamental problems, there are many avenues of research that aim to push the boundaries of current computer vision systems, including perception of visual input beyond the visible spectrum of light \cite{teutsch2022computer}, multi-agent and multi-view integration \cite{wang2023core} and understanding of underlying characteristics of physics, functionality and causality \cite{zhu2018visual}. Similarly, linguistic perception (in the field of natural language processing) has its own set of open problems that limit current AI systems, including reasoning abilities \cite{valmeekam2022large}, ability to handle arithmetic representations \cite{mandal2019solving}, and understanding nuanced cross-cultural linguistic phenomena such as euphemisms, idioms, and negations \cite{liu2023multilingual}.


\subsection{Contributions of the paper}

The primary goal of this paper is to review AI techniques, theories and methods related to perception, that share characteristics with cognitive science, directly or indirectly. We begin by studying prevalent theories from cognitive science. Cognitive science is a multi-disciplinary field consisting of many sub-disciplines related to functions of the brain. Of the various sub-disciplines, we majorly focus on three sub-disciplines of cognitive science -- \textbf{neuroscience} (the study of the brain), \textbf{psychology} (the study of the mind) and \textbf{linguistics} (the study of language and thought organization) \footnote{Cognitive science is widely defined as an culmination of five natural sciences -- psychology, neuroscience, linguistics, anthropology and philosophy; and their intersection with computational intelligence \cite{thagard2013cognitive}. In this discussion, we do not cover the fields of philosophy and anthropology, due to their limited contribution towards the current understanding of cognitive functions and their underlying mechanisms. However, it should also be kept in mind that, while philosophy and anthropology are not relevant to this particular discussion, they may be immensely helpful in other forms of studies, such as modeling of social behavior and ethics in AI \cite{becker2021anthropology, mccarthy1995has}.}. 


From an academic perspective, cognitive science and AI are largely treated as independent fields of study \cite{contreras2022quantifying}. The last few decades have seen limited interdisciplinary efforts, leading to distinct forms of formulating problem statements and theoretical assumptions \cite{dale2009explanatory}, modeling, and methods of application. 
This also leads to a knowledge gap, where AI practitioners lack understanding of cognitive science, and conversely, cognitive scientists lack awareness about the trends in AI. 
Hence,  translation of concepts from cognitive science to problem-specific AI research
involves a degree of resistance for researchers. This paper aims to bridge this gap by providing a fresh approach on cognitive AI through a bottom-up approach, which facilitates the merger of fundamental concepts from cognitive science  with the current prevalent methods in AI,  including that of data modeling, model training and architectural designing.

The main contributions of this paper are as follows. 
\begin{enumerate}

\item \textbf{Comprehensive state of cognitive AI}: Firstly, we provide a fresh perspective on cognitive science inspired AI, by studying AI methods and techniques that intentionally or indirectly share characteristics with cognitive science, while staying aligned with current research trends. This paper, hence, serves as a collection of various methods at the disposal of researchers to build effective cognitively inspired AI models.

    \item \textbf{Comprehensive state of cognitive science}: Secondly, this paper studies the current state of cognitive science from a contemporary perspective.
    Since cognitive science is an extremely wide field of research, we select theories through the criteria that the theories must be widely accepted and relevant (i.e., they must be cited by a large number of academic papers, especially in recent works), must not be in conflict with an equally popular theory, and must provide an interesting perspective to the field of AI, such as the prospect of better performance or computational efficiency.

    \item \textbf{Gaps in AI research}: Finally, through this process, we point out the obvious gaps in AI research, which can be addressed by studying principles and theories from cognitive science. Thus, this paper presents many potential directions of research in AI, that can be addressed through inspiration from cognitive science. 
\end{enumerate}

\section{\textbf{Background}} \label{sec: the perceptual pathway}
    
 

\subsection{History of cognitive AI}

 The history of cognitive based AI modeling dates as back as the history of AI research itself. Primitive developments in artificial neural networks and learning algorithms were, in essence, a crude replication of the human brain \cite{kycia2023biological}. Earlier forms of cognitive frameworks for AI included theoretical and philosophical propositions of the working of the human brain. These included theories such as the Global Workspace Theory \cite{baars2005global}, the Society of Mind \cite{minsky1987society} and the Pandemomium Theory \cite{bjork2018selfridge}, which essentially formulated cognitive functions and emotions as a collection of numerous underlying small models, the combination of which results in sophisticated models. Other forms of philosophical cognitive frameworks included the CogAff framework \cite{sloman2002cogaff}, which separates different cognitive processes into distinct layers, organized in a structure that involves both vertical hierarchy and parallelization. However, these frameworks were primarily conceptual in nature, and thus had limited practical applications. These frameworks inspired the next generation of cognitive architectures, which defined cognitively inspired layers of an AI system more definitively. These cognitive architectures, such as SOAR \cite{laird2019soar} and ACT-R \cite{ritter2019act}, majorly focus on modeling high-level cognitive functions like memory, learning and attention,  
  in a symbolic fashion. With the rise of advanced deep learning methods, various hybrid architectures, which combined symbolic reasoning with data-driven learning are becoming increasingly popular \cite{goertzel2013cogprime, madl2016towards}.  

Cognitive architectures provide valuable blueprints for defining relations between various cognitive functions from a broad perspective. However, a detailed review on the cognitive architectures of the past few decades \cite{kotseruba202040} revealed that, while a significant number of cognitive architectures had some form of a cognitive-based layout of various functions, the said functions have little to no cognitive influence in their core mechanism. Notably, perception has often been merely treated as a mechanism to capture signals electronically, where signals are more or less represented as standard discrete input to subsequent functions. Without  levying the advantage of cognitive sciences at the foundational layers of these architectures, they are bound to become bulky and inefficient. This inefficiency stems from  connecting various modules, each limited by their own technological constraints. 

\subsection{Perception in the brain}
\begin{figure}
    \centering
    \includegraphics[width = \columnwidth]{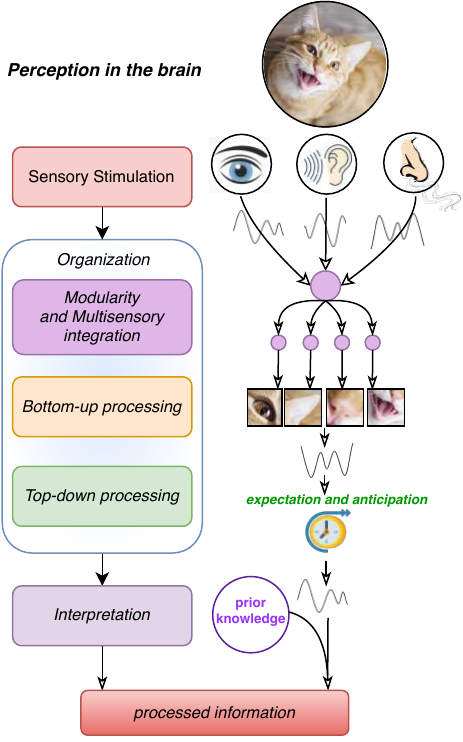}
    \caption{Processes of perception in the brain. Perception is the collective outcome of five major processes -- sensory stimulation, modularity and multisensory integration, bottom-up processing, top-down processing and interpretation. For each of these processes, we study concepts from neuroscience, psychology and linguistics, and draw parallels with AI.}
    \label{fig:processes_of_perception}
\end{figure}

We start by taking a deeper look into the meaning of perception in the brain, and breaking down the process of perception into its various sub-processes. Perception is the interpretation of the environment by an agent (biological or artificial) through sensory input. This process involves the reception and processing of raw information gathered from the surroundings. However, the interpretation of this raw information is not done on its own; it is influenced by various factors and undergoes several stages before leading to meaningful understanding and action. The human brain aggregates perceptual information from various sensory sources, such as  vision, audition and olfaction,  all of which are processed and stored as neuro-electric signals~\cite{paller2004electrical}.  Perception consists of three major processes -- \textbf{sensory stimulation} (or the identification of sensory input), \textbf{organization} of information from different sources~\cite{sarkar2003introduction} and \textbf{interpretation} of processed neural representations~\cite{schacter2011psychology}. These three sequential processes involve various sub-processes that are crucial for efficient information processes. Understanding the organization, sequence and relation between these different processes is crucial for capable AI models to be developed.  
\begin{enumerate}
    \item \textbf{\textit{Sensory stimulation}}~\cite{goldstein2016sensation}: The first step in the perception process is to take in a sensory input signal, which contains various different forms of signals (such as visual and auditory).
    
    \item \textit{\textbf{Organization}}: The sensory inputs are then processed in various stages in specialized perceptual pathways.
    \begin{enumerate}
         
        \item \textbf{Modularity and multisensory integration}~\cite{deroy2015modularity}: Features are hierarchically processed in different modules or areas of the brain, giving rise to highly specialized functionality, all of which are integrated together to form a common concept. Different modalities, such as vision, audition, and touch consist of complementary information and are integrated together at various stages, hence improving the robustness of perception. 

       \item \textbf{{Bottom-up processing}}~\cite{rauss2013bottom}: The sensory input is passed through various stages of processing in a bottom-up fashion, where features (attributes) are extracted and combined to form complex concepts, relations and patterns.
        
        \item \textbf{Top-down processing}~\cite{theeuwes2010top, dijkstra2017distinct}: The perceptual pathway is not a fixed collection of processes but is dynamically guided by other high-level cognitive processes, such as attention and memory, in a ~top-down fashion of processing information.

    \end{enumerate}
    \item \textbf{\textit{Interpretation}}~\cite{kording2007causal}: Processed sensory information is then combined with prior knowledge and cues through prediction and inference to give more informed results.
\end{enumerate}

These processes can be visualized in detail through Figure \ref{fig:processes_of_perception}. In the subsequent sections, we explore each one of these five processes (sensory stimulation, modularity and multisensory integration, bottom-up processing, top-down processing and interpretation) of perception individually. For each process, we gather relevant and prevalent theories from the domains of neuroscience, psychology and linguistics, where we study various theories and principles, and draw parallels with AI -- highlighting the current state-of-the-art and gaps in research that can potentially be addressed through following the corresponding cognitive principles.

\begin{table*}[]

\resizebox{\textwidth}{!}{

\begin{tabular}{llll}
\\ \hline
\rowcolor{green!10}
\multicolumn{4}{|c|}{\textbf{SENSORY STIMULATION}} \\ \hline
\rowcolor{lime!10}
\multicolumn{1}{|l|}{\textbf{Domain}} & \multicolumn{1}{l|}{\textbf{Topic}} & \multicolumn{1}{l|}{\textbf{Cognitive Theory}} & \multicolumn{1}{l|}{\textbf{Corresponding theory in AI}} \\ \hline

\rowcolor{orange!5}
\multicolumn{1}{|l|}{} & \multicolumn{1}{l|}{} & \multicolumn{1}{l|}{Retinotopicity \cite{saenz2011striking}} & \multicolumn{1}{l|}{Inherent spatial organization in CNNs \cite{mostafa2021visualizing}} \\ \cline{3-4} 

\rowcolor{orange!5}

\multicolumn{1}{|l|}{} & \multicolumn{1}{l|}{} & \multicolumn{1}{l|}{Cortical magnification \cite{born2015cortical}} & \multicolumn{1}{l|}{ \begin{tabular}[c]{@{}l@{}}Fovea-inspired cortical sampling \cite{da2021convnets, jansson2022scale} and data \\  transforms \cite{wang2021use} \end{tabular}} \\ \cline{3-4} 

\rowcolor{orange!5}
\multicolumn{1}{|l|}{} & \multicolumn{1}{l|}{\multirow{-3}{*}{Signal organization}} & \multicolumn{1}{l|}{Tonotopicity \cite{humphries2010tonotopic}} & \multicolumn{1}{l|}{Frequency selective capability in neural networks \cite{kell2018task}} \\ \cline{2-4} 

\rowcolor{orange!5}

\multicolumn{1}{|l|}{} & \multicolumn{1}{l|}{} & \multicolumn{1}{l|}{} & \multicolumn{1}{l|}{\begin{tabular}[c]{@{}l@{}}Chromaticity estimation methods for light, color and contrast\\  invariance \cite{land1977retinex,funt2010rehabilitation, van2005color,van2007edge,banic2017unsupervised, barnard2000improvements,finlayson2006gamut,van2007using,gijsenij2010color, finlayson2013corrected, gehler2008bayesian, chakrabarti2011color, banic2014color,agarwal2007machine,xiong2006estimating, cheng2015effective, qiu2020color, sidorov2020artificial, barron2015convolutional, yang2023learning }\end{tabular}} \\ \cline{4-4} 

\rowcolor{orange!5}
\multicolumn{1}{|l|}{} & \multicolumn{1}{l|}{} & \multicolumn{1}{l|}{} & \multicolumn{1}{l|}{Scale invariant architectures for size constancy \cite{xu2014scale, jansson2022scale}} \\ \cline{4-4}

\rowcolor{orange!5}
\multicolumn{1}{|l|}{} & \multicolumn{1}{l|}{} & \multicolumn{1}{l|}{\multirow{-3}{*}{Perceptual constancy \cite{cohen2015perceptual}}} & \multicolumn{1}{l|}{\begin{tabular}[c]{@{}l@{}}Estimation of perturbed view \cite{huang2022shape, agrawal2021learning} and constrastive learning   \\ methods \cite{gu2021staying} for shape invariance\end{tabular}} \\ \cline{3-4}

\rowcolor{orange!5}
\multicolumn{1}{|l|}{\multirow{-7}{*}
{\textbf{Neuroscience}}} & \multicolumn{1}{l|}{\multirow{-4}{*}{\begin{tabular}[c]{@{}l@{}}Perceptual constancy and\\ object permanence\end{tabular}}} & \multicolumn{1}{l|}{Object permanence \cite{bonzonpiaget}} & \multicolumn{1}{l|}{Object tracking methods \cite{shamsian2020learning, tokmakov2021learning, tokmakov2022object}} \\ \hline

\rowcolor{red!10}
\multicolumn{1}{|l|}{} & \multicolumn{1}{l|}{} & \multicolumn{1}{l|}{Sensitivity \cite{mcnicol2005primer}} & \multicolumn{1}{l|}{Encoder-decoder based signal denoising approaches \cite{ilesanmi2021methods, kashyap2021speech}} \\ \cline{3-4} 
\rowcolor{red!10}
\multicolumn{1}{|l|}{\multirow{-2}{*}{\textbf{Psychology}}} & \multicolumn{1}{l|}{\multirow{-2}{*}{\begin{tabular}[c]{@{}l@{}}Selective filtering of sensory  \\ signals \end{tabular}}} & \multicolumn{1}{l|}{Response criterion \cite{mcnicol2005primer}} & \multicolumn{1}{l|}{Input signal selectivity \cite{rabanser2022selective}} \\ \hline

\rowcolor{yellow!10}
\multicolumn{1}{|l|}{} & \multicolumn{1}{l|}{} & \multicolumn{1}{l|}{} & \multicolumn{1}{l|}{Predetermined tokenization methods \cite{mielke2021between, rai2021study,michelbacher2013multi, ma2022searching, song2020fast}} \\ \cline{4-4} 

\rowcolor{yellow!10}
\multicolumn{1}{|l|}{\multirow{-2}{*}{\textbf{Linguistics}}} & \multicolumn{1}{l|}{\multirow{-2}{*}{Phonological processing}} & \multicolumn{1}{l|}{\multirow{-2}{*}{Categorical perception \cite{harnad2003categorical, mompean2006phoneme}}} & \multicolumn{1}{l|}{Token grouping through dynamic learning \cite{tay2021charformer}} \\ \cline{2-4} 

\hline 



\\ \hline

 \rowcolor{green!10}
\multicolumn{4}{|c|}{\textbf{MODULARITY AND MULTISENSORY INTEGRATION}} \\ \hline

\rowcolor{lime!10}

\multicolumn{1}{|l|}{\textbf{Domain}} & \multicolumn{1}{l|}{\textbf{Topic}} & \multicolumn{1}{l|}{\textbf{Cognitive Theory}} & \multicolumn{1}{l|}{\textbf{Corresponding theory in AI}} \\ \hline

\rowcolor{orange!5}

\multicolumn{1}{|l|}{} & \multicolumn{1}{l|}{} & \multicolumn{1}{l|}{} & \multicolumn{1}{l|}{Explicit modular implementation in large parametric models \cite{ponti2021parameter}}  \\ \cline{4-4} 

\rowcolor{orange!5}
\multicolumn{1}{|l|}{} & \multicolumn{1}{l|}{\multirow{-2}{*}{\begin{tabular}[c]{@{}l@{}}Extraction of multimodal \\ features\end{tabular}}} & \multicolumn{1}{l|}{\multirow{-2}{*}{\begin{tabular}[c]{@{}l@{}}Hierarchical feature extraction\\ in specialized regions \cite{hollier1999multi}\end{tabular}}} & \multicolumn{1}{l|}{Natural modular organization of data in certain architectures \cite{zhang2023emergent}} \\ \cline{2-4} 

\rowcolor{orange!5}
\multicolumn{1}{|l|}{\multirow{-3}{*}{\textbf{Neuroscience}}} & \multicolumn{1}{l|}{Integration of modalities} & \multicolumn{1}{l|}{Temporal coincidence theory \cite{montague1994predictive}} & \multicolumn{1}{l|}{Nil} \\ \hline 


\rowcolor{red!10}
\multicolumn{1}{|l|}{} & \multicolumn{1}{l|}{\begin{tabular}[c]{@{}l@{}}Nature of modular \\ organization\end{tabular}} & \multicolumn{1}{l|}{\begin{tabular}[c]{@{}l@{}}Bergeron's theory of\\  modularity \cite{bergeron2008cognitive}\end{tabular}}  & 
\multicolumn{1}{l|}{Cross-modal learning  \cite{goyal2020cross, wang2016comprehensive}} 
\\ \cline{2-4} 

\rowcolor{red!10}
\multicolumn{1}{|l|}{} & \multicolumn{1}{l|}{} & \multicolumn{1}{l|}{} & 
 \multicolumn{1}{l|}{\begin{tabular}[c]{@{}l@{}}Classical multimodal feature fusion (late fusion, early fusion)\\  \cite{bayoudh2021survey, atrey2010multimodal}\end{tabular}}
\\ \cline{4-4} 

\rowcolor{red!10}
\multicolumn{1}{|l|}{} & \multicolumn{1}{l|}{} & \multicolumn{1}{l|}{} & \multicolumn{1}{l|}{\begin{tabular}[c]{@{}l@{}}Dynamic integration methods (attention-based  approaches \\ \cite{nagrani2021attention,qian2021multimodal} and mixture-of-experts models \cite{shazeer2017outrageously})\end{tabular}} \\ \cline{4-4} 

\rowcolor{red!10}
\multicolumn{1}{|l|}{\multirow{-4}{*}{\textbf{Psychology}}} & \multicolumn{1}{l|}{\multirow{-3}{*}{\begin{tabular}[c]{@{}l@{}}Functional flexibility\\ due to dynamic\\  integration\end{tabular}}} & \multicolumn{1}{l|}{\multirow{-3}{*}{Dynamical systems theory  \cite{favela2020dynamical}}} & \multicolumn{1}{l|}{Progressive integration of modalities\cite{xue2023dynamic} }\\ \hline

\rowcolor{yellow!10}

\multicolumn{1}{|l|}{} & \multicolumn{1}{l|}{} & \multicolumn{1}{l|}{} & \multicolumn{1}{l|}{Storing knowledge in parametric spaces \cite{petroni2019language, alkhamissi2022review}} \\ \cline{4-4} 

\rowcolor{yellow!10}
\multicolumn{1}{|l|}{\multirow{-2}{*}{\textbf{Linguistics}}} & \multicolumn{1}{l|}{\multirow{-2}{*}{Mental lexicon}} & \multicolumn{1}{l|}{\multirow{-2}{*}{Mental lexicon \cite{ullman2007biocognition, farahian2011mental}}} & \multicolumn{1}{l|}{Independent knowledge base entities \cite{mialon2023augmented, liu2022relational, wilmot2021memory}} \\ \hline \\
 

 \hline
\rowcolor{green!10}
\multicolumn{4}{|c|}{\textbf{BOTTOM-UP PROCESSING}} \\ \hline
\rowcolor{lime!10}
\multicolumn{1}{|l|}{\textbf{Domain}} & \multicolumn{1}{l|}{\textbf{Topic}} & \multicolumn{1}{l|}{\textbf{Cognitive Theory}} & \multicolumn{1}{l|}{\textbf{Corresponding theory in AI}} \\ \hline

\rowcolor{orange!10}
\multicolumn{1}{|l|}{} & \multicolumn{1}{l|}{} & \multicolumn{1}{l|}{Population encoding \cite{krahe2002stimulus}} & \multicolumn{1}{l|}{Adversarial training for robustness \cite{chen2022adversarial}} \\ \cline{3-4} 

\rowcolor{orange!10}
\multicolumn{1}{|l|}{\multirow{-2}{*}{\textbf{Neuroscience}}} & \multicolumn{1}{l|}{\multirow{-2}{*}{Signal propagation}} & \multicolumn{1}{l|}{Sparse encoding \cite{beyeler2019neural}} & \multicolumn{1}{l|}{Out-of-distribution detection for sensitivity \cite{wilson2023safe}} \\ \hline

\rowcolor{red!10}
\multicolumn{1}{|l|}{} & \multicolumn{1}{l|}{} & \multicolumn{1}{l|}{Feature detection theory \cite{pelli2006feature}} & \multicolumn{1}{l|}{Feature detection based pattern recognition \cite{lecun2015deep}} \\ \cline{3-4} 

\rowcolor{red!10}
\multicolumn{1}{|l|}{} & \multicolumn{1}{l|}{\multirow{-2}{*}{\begin{tabular}[c]{@{}l@{}}Pattern recognition \\ mechanism\end{tabular}}} & \multicolumn{1}{l|}{Geon theory \cite{biederman1993geon}} & \multicolumn{1}{l|}{Point cloud representation method \cite{qi2017pointnet, ran2022surface}} \\ \cline{2-4} 
\rowcolor{red!10}
\multicolumn{1}{|l|}{} & \multicolumn{1}{l|}{} & \multicolumn{1}{l|}{\multirow{2}{*}{Gestalt properties \cite{wertheimer1938gestalt}}} & \multicolumn{1}{l|}{\begin{tabular}[c]{@{}l@{}}Gestalt estimation methods in traditional machine  \\learning\cite{kanizsa1981completamento, desolneux2004gestalt, amanatiadis2018understanding}\end{tabular}} \\ \cline{4-4} 
\rowcolor{red!10}
\multicolumn{1}{|l|}{\multirow{-4}{*}{\textbf{Psychology}}} & \multicolumn{1}{l|}{\multirow{-2}{*}{\begin{tabular}[c]{@{}l@{}}Gestalt properties of\\  visual perception\end{tabular}}} & \multicolumn{1}{l|}{} & \multicolumn{1}{l|}{\begin{tabular}[c]{@{}l@{}}Gestalt properties in neural networks   \cite{hoerhan2021gestalt,kim2008object, baker2018deep, funke2021five, kim2021neural}\end{tabular}}  \\ 

\hline

\rowcolor{yellow!10}

\multicolumn{1}{|l|}{} & \multicolumn{1}{l|}{\begin{tabular}[c]{@{}l@{}}Incremental conceptualization \\ and levels of abstractions\end{tabular}} & \multicolumn{1}{l|}{
\begin{tabular}[c]{@{}l@{}}Why-vs-how framework   \cite{freitas2004influence, spunt2016neural}\end{tabular}
} & \multicolumn{1}{l|}{\begin{tabular}[c]{@{}l@{}}Complex context-aware reasoning in deep LLM layers \cite{zhang2023wider}\end{tabular}}  \\ \cline{2-4} 
\rowcolor{yellow!10}

\multicolumn{1}{|l|}{} & \multicolumn{1}{l|}{} & \multicolumn{1}{l|}{} & \multicolumn{1}{l|}{\begin{tabular}[c]{@{}l@{}}Exploratory creativity methods \cite{franceschelli2021creativity}\\ (e.g. reinforcement learning \cite{cohen2007reinforcement})\end{tabular}} \\ \cline{4-4} 
\rowcolor{yellow!10}

\multicolumn{1}{|l|}{} & \multicolumn{1}{l|}{} & \multicolumn{1}{l|}{\multirow{-2}{*}{\begin{tabular}[c]{@{}l@{}}Creative engine in the\\ default mode network \cite{herrmann1991creative}\end{tabular}}} & \multicolumn{1}{l|}{\begin{tabular}[c]{@{}l@{}}Combinatorial creativity methods \cite{franceschelli2021creativity} (e.g.\\ Probabilistic time modeling \cite{neumann2019future})\end{tabular}} \\ \cline{3-4} 
\rowcolor{yellow!10}

\multicolumn{1}{|l|}{\multirow{-4}{*}{\textbf{Linguistics}}} & \multicolumn{1}{l|}{\multirow{-3}{*}{Creativity}} & \multicolumn{1}{l|}{Spreading activation theory \cite{collins1975spreading}} & \multicolumn{1}{l|}{\begin{tabular}[c]{@{}l@{}}Transformation creativity methods\cite{franceschelli2021creativity}(e.g. diffusion \\  models \cite{croitoru2023diffusion},   adversarial networks \cite{li2022comprehensive})\end{tabular}} \\ \hline \\


 \hline
 \rowcolor{green!10}
\multicolumn{4}{|c|}{\textbf{TOP-DOWN PROCESSING}} \\ \hline
\rowcolor{lime!10}
\multicolumn{1}{|l|}{\textbf{Domain}} & \multicolumn{1}{l|}{\textbf{Topic}} & \multicolumn{1}{l|}{\textbf{Cognitive Theory}} & \multicolumn{1}{l|}{\textbf{Corresponding theory in AI}} \\ \hline

\rowcolor{orange!10}

\multicolumn{1}{|l|}{\textbf{Neuroscience}} & \multicolumn{1}{l|}{\begin{tabular}[c]{@{}l@{}}Neuronal synchronization \\ as a guiding mechanism\end{tabular}} & \multicolumn{1}{l|}{\begin{tabular}[c]{@{}l@{}}Neuronal synchronization theory \\ \cite{guevara2017neural}\end{tabular}} & \multicolumn{1}{l|}{Nil} \\ \hline

\rowcolor{red!10}
\multicolumn{1}{|l|}{\textbf{Psychology}} & \multicolumn{1}{l|}{\begin{tabular}[c]{@{}l@{}}Templates for guiding \\ bottom-up processing\end{tabular}} & \multicolumn{1}{l|}{Schema theory \cite{mcvee2005schema}} & \multicolumn{1}{l|}{Concept representation framework \cite{chang2021concept}} \\ \hline

\rowcolor{yellow!10}

\multicolumn{1}{|l|}{} & \multicolumn{1}{l|}{} & \multicolumn{1}{l|}{Garden path theory \cite{frazier1982making}} & \multicolumn{1}{l|}{\multirow{2}{*}{\begin{tabular}[c]{@{}l@{}}Transformer based language model families (can exhibit both  \\ garden path \cite{jurayj2022garden, irwin2023bert}  and constrain based \\ characteristics \cite{garbacea2022constrained, davis2021uncovering} in different task settings)\end{tabular}}} \\ \cline{3-3}
\rowcolor{yellow!10}

\multicolumn{1}{|l|}{\multirow{-2}{*}{\textbf{Linguistics}}} & \multicolumn{1}{l|}{\multirow{-2}{*}{\begin{tabular}[c]{@{}l@{}}Filtering competing \\ activation\end{tabular}}} & \multicolumn{1}{l|}{\begin{tabular}[c]{@{}l@{}}
     Constrain based theory\\ \cite{thompson1999effects}
\end{tabular}} & \multicolumn{1}{l|}{} \\ \hline \\

 \hline

 \rowcolor{green!10}
\multicolumn{4}{|c|}{\textbf{INTERPRETATION}} \\ \hline
\rowcolor{lime!10}
\multicolumn{1}{|l|}{\textbf{Domain}} & \multicolumn{1}{l|}{\textbf{Topic}} & \multicolumn{1}{l|}{\textbf{Cognitive Theory}} & \multicolumn{1}{l|}{\textbf{Corresponding theory in AI}} \\ \hline

\rowcolor{orange!5}

\multicolumn{1}{|l|}{} & \multicolumn{1}{l|}{} & \multicolumn{1}{l|}{} & \multicolumn{1}{l|}{\begin{tabular}[c]{@{}l@{}}Hierarchical propagation of error signals\\ through predictive coding principles \cite{mikulasch2023error}\end{tabular}} \\ \cline{4-4} 
\rowcolor{orange!5}

\multicolumn{1}{|l|}{} & \multicolumn{1}{l|}{} & \multicolumn{1}{l|}{\multirow{-2}{*}{\begin{tabular}[c]{@{}l@{}}Predictive coding theory\\  \cite{shipp2016neural, caucheteux2023evidence}\end{tabular}}} & \multicolumn{1}{l|}{\begin{tabular}[c]{@{}l@{}}Recursiveness \cite{salehinejad2017recent} and bidirectionalism \cite{graves2005bidirectional} connectionist\\ models\end{tabular}} \\ \cline{3-4} 
\rowcolor{orange!5}

\multicolumn{1}{|l|}{} & \multicolumn{1}{l|}{} & \multicolumn{1}{l|}{} & \multicolumn{1}{l|}{Dopamine-based reinforcement learning \cite{richards2019deep}} \\ \cline{4-4} 
\rowcolor{orange!5}

\multicolumn{1}{|l|}{} & \multicolumn{1}{l|}{\multirow{-4}{*}{Predictive processes}} & \multicolumn{1}{l|}{\multirow{-2}{*}{Prediction error \cite{den2012prediction}}} & \multicolumn{1}{l|}{Spiking errors in spiking neural networks \cite{ghosh2009spiking}} \\ \cline{2-4} 
\rowcolor{orange!5}

\multicolumn{1}{|l|}{} & \multicolumn{1}{l|}{} & \multicolumn{1}{l|}{\begin{tabular}[c]{@{}l@{}}Variational free energy \\ optimization \cite{friston2007variational}\end{tabular}} & \multicolumn{1}{l|}{\begin{tabular}[c]{@{}l@{}}Out-of-distribution detection    \cite{yang2021generalized} and few-shot\\ learning methods \cite{lu2020learning}\end{tabular}} \\ \cline{3-4} 
\rowcolor{orange!5}

\multicolumn{1}{|l|}{\multirow{-6}{*}{\textbf{Neuroscience}}} & \multicolumn{1}{l|}{\multirow{-2}{*}{\begin{tabular}[c]{@{}l@{}}Optimization principles of\\ interpretation\end{tabular}}} & \multicolumn{1}{l|}{\begin{tabular}[c]{@{}l@{}}Expected free energy\\ optimization \cite{millidge2021whence}\end{tabular}} & \multicolumn{1}{l|}{\begin{tabular}[c]{@{}l@{}}Future event prediction through bayesian prediction \cite{millidge2021whence} and  \\ reinforcement learning \cite{ogishima2020combining}\end{tabular}} \\ \hline

\rowcolor{red!10}
\multicolumn{1}{|l|}{} & \multicolumn{1}{l|}{Nature of prediction} & \multicolumn{1}{l|}{Dual theory process \cite{gawronski2013dual}} & \multicolumn{1}{l|}{Estimation methods for heuristical prediction \cite{raiman1990order,mavrovouniotis1987reasoning,yip1996model, dubois1990order, de1984qualitative}} \\ \cline{2-4} 
\rowcolor{red!10}

\multicolumn{1}{|l|}{\multirow{-2}{*}{\textbf{Psychology}}} & \multicolumn{1}{l|}{Inference} & \multicolumn{1}{l|}{\begin{tabular}[c]{@{}l@{}}Prior-informed probabilistic\\ inference \cite{shams2022bayesian, aitchison2017or}\end{tabular}} & \multicolumn{1}{l|}{Bayesian inference methods \cite{wang2020survey, wilson2020case}} \\ \hline

\rowcolor{yellow!10}
\multicolumn{1}{|l|}{\textbf{Linguistics}} & \multicolumn{1}{l|}{Semantic understanding} & \multicolumn{1}{l|}{\begin{tabular}[c]{@{}l@{}}Discourse representation theory \cite{kamp2011discourse}\end{tabular}} & \multicolumn{1}{l|}{Coreference relation tracking in language models \cite{tenney2019bert, clark2019does}}  \\ \hline
    
    \end{tabular}
    } 

\vspace{10pt}
\caption{Summary of methods in AI that derive inspiration from cognitive science. 
}
\label{tab: review}
\end{table*}

 \section{\textbf{Sensory stimulation}}  \label{sensory stimulation}

In this section, we highlight various properties of the initial stages of perception in the brain. Specifically, we highlight various interesting characteristics of the visual pathway, such as the robustness of the retina to spatial distances, and the of the linguistic pathway, such as hearing a continuous speech as distinct sounds for efficient understanding. The findings are highlighted in Figure \ref{fig:topics_covered} and Table \ref{tab: review}.
 
\subsection{\textbf{In neuroscience}}


\subsubsection{\textbf{Signal organization}} 


The neurons in different sensory pathways are organized systematically, preserving a structured spatio-temporal input mapping. For example, in the visual pathway, the signals originating from retinal cells are arranged to form a spatial map \textit{(retinotopicity)} \cite{saenz2011striking}, such that neuronal distances are correlated with geometrical distances in the visual receptive field of the eye. Similarly, in the auditory pathway, the cochlear neurons are arranged \textit{tonotopically} in the auditory pathway \cite{humphries2010tonotopic} which provides a structured frequency gradient map, allowing the brain to distinguish frequencies efficiently and even localize the source of a sound.  

Retinotopicity exhibits some interesting properties, such as 
\textit{cortical magnification}, which allows efficient processing of high-resolution visual input \cite{born2015cortical}. This  is characterized with a neural density gradient around the retinotopic center, resulting in high neural density at the center, and gradually decreasing neural density in the periphery. 

\underline{Parallel with AI:} Similar to \textit{retintopic properties} of the visual pathway, convolutional neural networks (CNNs) also demonstrate spatial organizational 
\cite{mostafa2021visualizing}. However, convolutional operations, by nature, lack the properties of \textit{cortical magnification} \cite{da2021convnets}. Convolutional operations, instead, sample the visual field uniformly, which leads to inadequate allocation of computational resources for visual areas requiring focussed processing. An attempt to implement such a fovea-inspired cortical sampling was proposed by \cite{da2021convnets}, where a simple neural layer was introduced on top of a CNN, resulting in many retinotopic organization properties including cortical magnification. In general, cortical magnification inspired data transforms have been shown to improve the performance of computer vision models \cite{wang2021use}.

Deep learning models aimed towards auditory tasks exhibit \textit{tonotopic} (frequency selective) properties as well. Certain layers of deep learning models are capable of signal discrimination when trained on constrained tasks \cite{kell2018task}. Even though there is no universal rule of selecting the tonotopic regions of a neural network, developing heuristics to identify task-specific frequency-selective regions in a given model may be fruitful. Analyzing the frequency response of neural networks can contribute towards many tonotopic abilities such as frequency discrimination, sound segregation \cite{micheyl2013auditory} and spatial localization \cite{francl2022deep}.

\subsubsection{\textbf{Perceptual constancy and object permanence}}

Visual transduction is characterized by \textit{perceptual constancy} \cite{cohen2015perceptual} and \textit{object permanence} \cite{bonzonpiaget}. Perceptual constancy is the ability to perceive objects in the visual field as static, despite physical variations such as lighting, color and contrast \cite{nunez2018cortical,morshedian2017light}  (carried out at the retinal level), and other variations such as size, shape, and warping/distortion \cite{hatfield2014psychological, wenderoth2006testing} (carried out in later stages of visual processing). Object permanence, on the other hand, is the ability to perceive the presence and state of an object even after its partial occlusion, or complete disappearance from the visual field. 

Both these mechanisms are related, and arise out of the synaptic ability to access representations of objects with only partial sensory input activation, and retain this representation even after the visual ceases to persist \cite{baird2002frontal}. This indicates towards a neural mapping mechanism that maps sparse sensory signals to higher dimensional information-rich representations in the initial stages of perception.

\underline{Parallel with AI}
In traditional computer vision modeling, \textit{light, color and contrast constancy} (i.e. robustness towards physical variations) is achieved through augmentation \cite{shorten2019survey}. However, this approach easily breaks down on out-of-distribution data \cite{afifi2019else}. An interesting alternative approach, referred to as  \textit{chromaticity estimation}, aims to regulate the effect of physical variations by estimating and correcting the color map of the visual input. 
Various types of  chromaticity estimation modeling approaches can be found in existing AI literature, such as (in increasing order of computational complexity) --

\begin{itemize}
    \item Statistical approaches (including distribution analysis methods \cite{land1977retinex,funt2010rehabilitation} and edge/filter based methods \cite{van2005color,van2007edge}).
    \item Traditional machine learning-based approaches (including unsupervised clustering methods 
 \cite{banic2017unsupervised}, pixel mapping methods \cite{barnard2000improvements,finlayson2006gamut,van2007using,gijsenij2010color, finlayson2013corrected}, probabilistic methods \cite{gehler2008bayesian, chakrabarti2011color}, regression methods \cite{banic2014color,agarwal2007machine,xiong2006estimating}, and decision-tree based methods \cite{cheng2015effective}).
 \item  Deep learning based approaches \cite{qiu2020color, sidorov2020artificial, barron2015convolutional}.
\end{itemize}

Selection between these three approaches is a trade-off between computational complexity and performance. Simpler statistical approaches may not be suitable for highly accurate perceptual constancy  but may be useful to simple AI perception systems in a computational-friendly manner. On the other hand, sophisticated approaches like deep learning methods are resource-intensive but can handle complex distributions and can perform light enhancement, exposure correction,  tone mapping, and noise suppression simultaneously \cite{yang2023learning}.

Previous studies have attempted to achieve \textit{size constancy}  through scale-invariant architectural approaches in CNN \cite{xu2014scale}. However, this approach does not generalize well to scales outside the training data distribution, because of the inability of convolutional models to process global-scale features \cite{baker2018deepglobal}. However, \cite{jansson2022scale} showed that a fovea-inspired convolutional architecture can perform on previously unseen scales efficiently. \textit{Shape constancy} methods, which refers to the invariance of models to perturbations in the viewpoint of the object (as in rotation, or warping), involve two types of approaches -- estimation of the perturbed view \cite{huang2022shape, agrawal2021learning}, and contrastive learning approaches, involving architectures that simultaneously learn various shape variations of an object \cite{gu2021staying}. 

\textit{Object permanence} has been an active area of research in the domain of object tracking, and is majorly driven by data-centric approaches \cite{shamsian2020learning, tokmakov2021learning, tokmakov2022object}. However, these solutions perform poorly and are not generalizable. 

Overall, it is observed that the current state of perceptual constancy and  permanence in AI is largely  reliant on the training distribution of the data provided to the models. These models perform poorly when partial sensory signals are provided to them. A higher dimension mapping strategy, similar to the brain, may be a better alternative to the current data-driven approach. 


\subsection{\textbf{In psychology}}

\subsubsection{\textbf{Selective filtering of sensory signals}}

Sensory stimulation is associated with interesting properties such as the ability to ignore noise and the preference for certain types of signals. According to the \textit{signal detection theory} \cite{mcnicol2005primer}, there are two major criteria used by the brain to filter the sensory input signal -- \textit{sensitivity} and \textit{response criterion} \cite{mcnicol2005primer}. 

Sensitivity refers to the ability to filter out true signals from a noisy environment. Response criterion, on the other hand, suggests that signals with strength only above a specific threshold are processed. Together, these act as a selective function that allows only the relevant and meaningful signals to be processed. The sensitivity criterion and response thresholds are affected by various factors such as expectation (anticipation) and prior knowledge and are context-specific in nature \cite{rahnev2011prior}. 

\underline{Parallel with AI}
Similar to \textit{sensitivity} in the signal detection theory, which relates to filtering noise from input, various \textit{signal denoising} approaches have been proposed using encoder-decoder architectural approaches \cite{ilesanmi2021methods, kashyap2021speech}. However, unlike the brain, where feedback connections guide the magnitude of signal sensitivity based on cues such as anticipation of a signal, predominant AI approaches are purely feed-forward. While some early work on attention-guided signal filtering methods provides foundational work for cue-driven denoising \cite{nguyen2021use,tian2020attention}, there is a lack of research on context-driven denoising based on anticipation of future signals. 

There has been some work on input signal \textit{selectivity} (similar to response criterion in signal detection theory) \cite{rabanser2022selective}, which aims to reject poor quality input signals. However, another important goal of the response criterion is to reject irrelevant signals in the agent's surrounding. 
Most deep learning models are trained in task-specific settings, which assumes that all input is relevant and clean. This can however be rectified with minor changes in the training regime,by training the model on  mixture of relevant and irrelevant signals, optimizing for task-relevant signals \cite{mannil2011rejection}.

\subsection{\textbf{In linguistics}}
   
\subsubsection{\textbf{Phonological processing}}
The first step of speech processing in the language processing pathway involves breaking down signals into simple and distinct units of sounds, known as phonemes in the primary auditory cortex \cite{mesgarani2008phoneme}. According to the \textit{categorical perception theory} \cite{harnad2003categorical, mompean2006phoneme}, phonemes are treated as independent categories, which are categorized based on efficient phenome recognition \cite{van2003efficient}.

\underline{Parallel with AI}
The major themes of phonological processing can be seen in fundamental techniques used in natural language processing (NLP) tasks in the field of AI. Similar to phoneme categorization, which involves breaking down incoming language signals into small chunks, are also use in NLP tasks  in the form of various tokenization techniques \cite{mielke2021between, rai2021study,michelbacher2013multi}. Many tokenization techniques such as subword tokenization \cite{ma2022searching} and wordpiece tokenization \cite{song2020fast} also emulate the efficient grouping of syllables as seen in phonological processing. This type of grouping ensures that groups of symbols that occur together more often are assigned an independent phenome category. Over time, the brain updates phenome categories (for example, when learning new languages, or during early development stages). Traditionally, NLP tasks use a fixed token vocabulary, with no ability to dynamically regroup token characters for efficient processing. Some works have attempted to learn character grouping techniques using a gradient-based approach \cite{tay2021charformer}, allowing efficient tokenization adapted to the language task at hand. However, updating the token vocabulary requires models to be retrained from scratch, unlike the brain, where phoneme recategorization can occur in parallel to linguistic processing, without any degrading effect on the signal quality.

\subsubsection{\textbf{Phonological buffer}}

According to \textit{Baddeley's working memory model} \cite{baddeley1999working}, the brain stores raw signals temporarily in a memory element known as \textit{phonological buffer} during language comprehension. The buffer helps recall recently heard phrases without significant neural processing, correct minor errors in the sensory input (such as wrongly pronounced words, or missing sounds) \cite{caramazza1986role}. It also maintains mental intermediate states, while allows efficient mental arithmetic and prediction of future events \cite{imbo2007role}. 

\underline{Parallel with AI}
Some characteristics of the phonological buffer are also seen in the form of hidden states in recurrent models \cite{ming2017understanding} (such as RNNs, GRUs, and LSTMs) and the self-attention mechanism in transformer-based architectures \cite{kim2021rethinking}. These hidden states store information about previous words and are updated recursively in each iteration. While some features of the phonological buffer are satisfied with this approach, such as deriving contextual information from previous words, some important properties of the phonological buffer are still not developed in AI models, such as the processing of low resource language tasks by bypassing the feedforward processing pipeline.

 \section{\textbf{Modularity and multisensory integration}}
 Modularity in the brain is a highly sophisticated process, where signals are incrementally divided and directed to different functional parts of the brain. Signal modularization (that divides signals to perform different processes independently) and signal integration (collection of the individually processed sub-signals to obtain a unified representation of information) go hand-in-hand in a highly structured yet dynamical method. A summary of these processes in the brain can be visualized in Figure \ref{fig:topics_covered} and Table \ref{tab: review}.

\subsection{\textbf{In neuroscience}}

\subsubsection{\textbf{Extraction of multimodal features}}


The human brain exhibits sophisticated modularity, through the division of perceptual processing into various specialized regions \cite{hollier1999multi}. Different modalities apply specific filters to the relevant sensory signal in hierarchical stages, thus extracting features in progressively increasing complexities.

\underline{Parallel with AI:} Over-parameterization of models (in the order of trillions of parameters in recent architectural approaches)~\cite{fedus2022switch} has demonstrated the ability of recent deep learning models to process a vast range of tasks simultaneously, without any explicit separation of signals. However, a modular architecture with separated units of computation for different tasks and modalities can provide many benefits, including parameter efficiency, prevention of signal interference between different tasks, and generalization to new emergent properties \cite{ponti2021parameter}. Such a modular approach in AI modeling has only been explored in limited directions. 

Modularity need not be explicitly designed architecturally. Certain architectures like the transformer  naturally result in modular groups of computation when trained on large general-purpose datasets \cite{zhang2023emergent}. However, identifying and adapting to the specific nature of different modules within a network can lead to improved overall performance. 

\subsubsection{\textbf{Integration of modalities}}

There exist specialized regions that are responsible for the integration of the processed sensory information from the various modalities \cite{angelaki2009multisensory}. An interesting theory related to perceptual integration is the \textit{temporal coincidence theory} \cite{montague1994predictive}, which suggests that temporally aligned neural features are integrated together. Temporal frequency matching allows unique patterns of intra-modal and cross-modal integration, allowing only relevant information to be combined.

\underline{Parallel with AI:} Majorly, multisensory integration is realized in AI by processing different sensory signals (such as vision, audio, and text) independently~\cite{gao2020survey} in a static manner, where a fixed architecture treats each modality identically. This approach neglects the different computational needs of different modalities~\cite{xue2023dynamic}. In contrast, the brain associate different types of neural features with unique frequencies (as proposed in the \textit{temporal coincidence theory}~\cite{montague1994predictive}), allowing only similar signals to associate. This dynamic grouping of signals can lead to improved performance by virtue of shared processing of different modalities (rather than independent processing). However, this concept has not been explored well in AI yet.

\subsection{\textbf{In psychology}}

\subsubsection{\textbf{Nature of modular organization}}




Unlike older theories that promote the isolation of cognitive modules in the brain~\cite{fodor1983modularity,baars1996understanding}, contemporary theories in cognitive sciences adopt a more unified approach to modular organization. \textit{Bergeron's theory of modularity} \cite{bergeron2008cognitive} posits a more flexible approach, where cognitive functions are viewed as domain-specific, but not modular in a strictly physical sense. Each high-level cognitive function, while fairly independent, consists of various lower-level functions (sub-modules), which can be shared by other cognitive functions. 

\underline{Parallel with AI:}
The existence of lower-level modules collectively shared by, and progressively unified to isolated higher-level modules (as proposed by Bergeron's theory of modularity) remains majorly unexplored in AI. While cross-modal learning \cite{goyal2020cross, wang2016comprehensive} presents some glimpses of sharing features across modalities in intermediate layers of a network, no major work explores progressive modularization of the model. Exploring this direction is an interesting opportunity for AI research, and is expected to enhance AI model performance and efficiency. 

\subsubsection{\textbf{Functional flexibility due to dynamic integration}}
Sharing of cognitive resources across modules gives rise to new emergent functions \cite{barsalou2008grounded}. This is formally supported by the \textit{dynamical systems theory} \cite{favela2020dynamical}, which claims that the integration of multisensory information takes place dynamically through the interaction of various perceptual components in arbitrary stages. Hence, multi-modal sensory information can be efficiently combined in a flexible manner.

\underline{Parallel with AI:} Classical multimodal models treated feature fusion in two major ways -- late fusion~\cite{bayoudh2021survey}, where modalities were treated completely independently, and early fusion, where different features are combined and normalized before any processing~\cite{atrey2010multimodal}. While both these approaches have their advantages and limitations, they share a common attribute of the lack of dynamic integration of modalities, as proposed by the \textit{dynamic systems theory}.

While some approaches provide crude solutions towards dynamic (i.e. task-specific learnable) integration of modalities (including attention-based approaches \cite{nagrani2021attention,qian2021multimodal} and mixture-of-expert models \cite{shazeer2017outrageously}), they still majorly integrate modalities over a single layer. A potentially viable solution is proposed in \cite{xue2023dynamic}, a progressive integration of modalities is proposed over multiple layers. The potential of such an approach can, however, only be realized by scaling this solution to models with large sizes of parameters and layers.

\subsection{\textbf{In linguistics}}

\subsubsection{\textbf{Mental lexicon}}

After phonological processing, speech signals are grouped together to form distinct words, and definitions are derived from a dictionary, known as the mental lexicon \cite{ullman2007biocognition}. 
This mental dictionary acts as a modular system connecting various related concepts, such as definitions, visual appearance,  spelling and pronunciation \cite{downie1985unit}. 

The mental lexicon has various interesting properties. For example, it is known to be hierarchically organized, where similar concepts are grouped together~\cite{farahian2011mental}, forming distinct clusters of shared representations.  Another interesting property is that the mental lexicon is highly dynamic, where new meanings are constantly learned and representations are constantly updated \cite{libben2022lexicon}, without affecting the efficiency of subsequent processes. 

\underline{Parallel with AI:} Unlike the concept of mental lexicon in the brain, which acts as the storage house of all conceptual knowledge in the brain, the mainstream approach in AI-based models is store information in parameters of the model implicitly \cite{petroni2019language}. This approach of using parameters as a knowledge base has significantly improved through research in the field of language modeling~\cite{alkhamissi2022review}. However, there are still some fundamental issues in this approach. As the size of the knowledge corpus increases, the models have to be scaled exponentially in terms of the number of parameters~\cite{kaplan2020scaling} to achieve the same theoretical performance. Larger models are also highly prone to incorrect language predictions~\cite{lazaridou2021mind}, and  updating knowledge is highly tricky because of the interdependent nature of parametric statistics with data distribution, unless the entire model is retrained, which can be expensive~\cite{alkhamissi2022review}.

Many works augment language models in AI to incorporate an independent knowledge base entity~\cite{mialon2023augmented}. This approach not only allows the language model to offload information retrieval to an external module with larger reliability, but also allows reduction of the model size in terms of parameters. 
The knowledge base entity may take many forms, such as a knowledge graph~\cite{liu2022relational} or a simpler look-up based method~\cite{wilmot2021memory}. One major advantage of an independent knowledge base entity is that it can be updated over time, without compromising the performance of the main prediction model. On the other hand, models are sensitive to the entity representations of knowledge base, which is why updating an knowledge base updating is tricky, and requires extensive analyses before an update.

\begin{figure*}
    \centering
    \includegraphics[width = \textwidth]{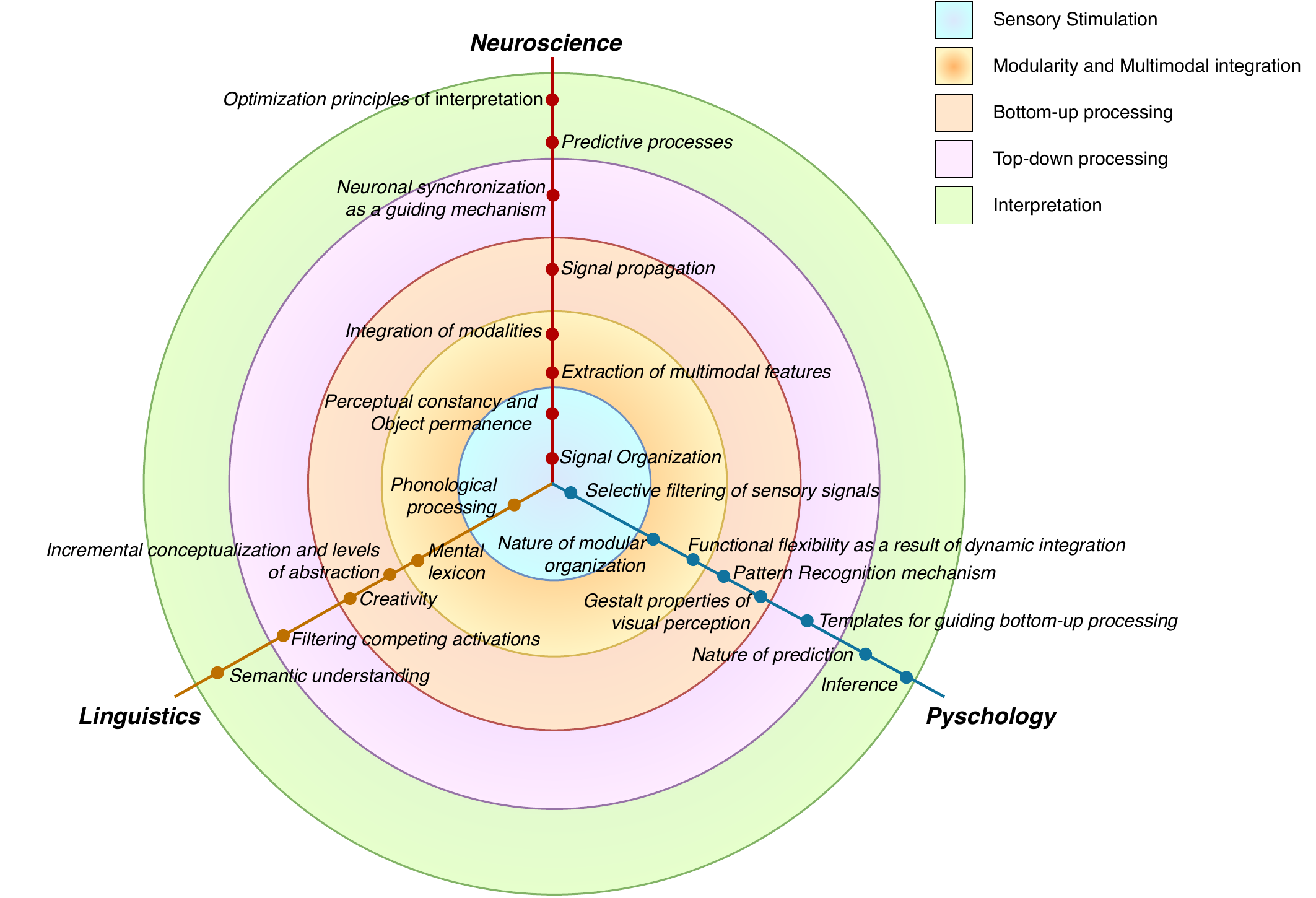}
    \caption{A visualization of all topics studied in this paper. We select various relevant theories of perception from the different sub-domains of cognitive sciences that have valuable insights to offer to AI.  }
    \label{fig:topics_covered}
\end{figure*}

 \section{\textbf{Bottom-up processing}}
Bottom-up processing, which is the process by which features are extracted from signals is a multi-stage process, where each stage (or level) is responsible for extraction of special types of features from sensory signals. Consequently, it is observed that the various stages of bottom-up processing have different functional mechanism. A brief summary of these methods, as understood by cognitive sciences, can be referred to in Figure \ref{fig:topics_covered} and Table \ref{tab: review}.

\subsection{\textbf{In neuroscience}}

\subsubsection{\textbf{Signal propagation}} 
At the core of bottom-up processing in the brain, the propagation of sensory signals through specific regions of the brain results in the extraction of meaningful features. The nature of information propagation, as suggested by neural encoding experiments \cite{rolls2011neuronal}, is highly statistical. Sensory information is typically propagated simultaneously by neurons within a neuronal group \cite{krahe2002stimulus} (known as \textit{population encoding}), ensuring  robustness and generalization of the signal by suppressing noisy information.
However, certain sensory signals are propagated through \textit{sparse encoding} \cite{beyeler2019neural}, which involves the propagation of signals through a relatively small number of neurons. This allows the detection of specific signals with high sensitivity.


\underline{Parallel with AI:}
While training AI models to achieve robustness (through techniques such as adversarial training \cite{chen2022adversarial}) allows models to withstand data perturbations, sensitivity is useful to detect out-of-distribution data \cite{wilson2023safe}. However, training models to achieve both robustness and sensitivity is tricky, since the two properties are contradictory. While the former optimizes model to ignore deviation from a statistical average, the latter promotes training to identify minor deviations with high precision.  However, some modeling techniques have been developed to strike a balance between the two. This includes stabilization of sensitive parameters \cite{zhang2020interpreting1} and the use of specific loss functions \cite{kim2021balancing}.

\subsection{\textbf{In psychology}}

\subsubsection{\textbf{Pattern recognition mechanism}}


Many theories have been proposed to explain pattern recognition properties of the brain 
 \cite{shugen2002framework}. However, the most recent and widely accepted theory of object recognition mechanisms is the \textit{recognition-by-component (RBC) theory}. RBC theory is a general family of theories, which argues that objects are treated as the collection of their constituent parts and properties. 

This includes the \textit{feature detection theory} \cite{pelli2006feature}, according to which recognition of visual stimuli occurs as a result of various feature identifiers in the brain, including edges, colors, shapes, and texture. Another theory in the RBC family is the \textit{geon theory} \cite{biederman1993geon}, which hypothesizes that object recognition is a result of the clustering of various small constituent 3D visual representations (\textit{geons}). 

\underline{Parallel with AI:}
Pattern recognition in AI is primarily developed on the lines of \textit{feature detection theory} \cite{lecun2015deep}. 
However, the \textit{geon theory} currently has limited applications in AI, such as in point cloud representation methods in 3D computer vision \cite{qi2017pointnet, ran2022surface}. Many characteristics of the geon theory such as breaking down complex objects into a limited number of simple shapes,  can be integrated into mainstream AI research. This can potentially result in efficient visual relationship detection \cite{cheng2022visual}, by virtue of reduced search spaces for AI models, leading to better relations and patterns between geons. It can also solve various perceptual constancy problems such as size and shape invariance. 


\subsubsection{\textbf{Gestalt properties of visual perception}}
An interesting set of organizational properties of visual bottom-up processing are the \textit{gestalt properties} \cite{wertheimer1938gestalt}.
This set of properties treats the perception process as a collective sensation of the environment, rather than individual processing of elements. The gestalt theory describes visual perception to have various properties such as proximity, similarity, continuity, closure, symmetry, convexity, pr$\ddot{a}$gnanz, parallelism, and others \cite{lahey1992psychology,palmer2002organizing,prinzmetal1995loosening}. 

\underline{Parallel with AI:}
There have been many attempts to produce generalized gestalt estimates through machine learning approaches \cite{kanizsa1981completamento, desolneux2004gestalt}. In general, it has been demonstrated that many AI models, including neural network based approaches are capable of demonstrating most gestalt effects in specific conditions \cite{amanatiadis2018understanding}.

Specific gestalt properties have been studied in depth in the field of deep learning. For example, simple neural network-based computer vision approaches have demonstrated similarity and proximity gestalts \cite{hoerhan2021gestalt,kim2008object}, as well as homogeneity \cite{yan2018unsupervised}. However, the property of closure has proven to be more difficult to achieve than others \cite{baker2018deep}, where only specific settings illicit closure properties \cite{funke2021five}. Recent studies argue that contour closure should not be considered as a physical property of a visual space, but rather an internal relation that guides visual perception \cite{kim2021neural}.

\subsection{\textbf{In linguistics}}

\subsubsection{\textbf{Incremental conceptualization and levels of abstractions}}

Raw definitions of word sequences are incrementally processed to derive semantic meaning from speech. From the mental lexicon, multiple related sub-concepts are derived and processed to form incremental levels of conceptual abstractions. According to the \textit{why-vs-how framework} \cite{freitas2004influence}, lower level abstractions, corresponding to ``how", represent the underlying functional characteristics of the concept, while higher level abstractions in the later stages of bottom-up processing, corresponding to ``why", represent semantic context of a thought. This framework also quantifies levels of abstractions into concrete conceptual parameters \cite{spunt2016neural}
, such as concreteness, specificity, imageability \cite{richardson1975concreteness} and neutrality of the intended semantic message. 

\underline{Parallel with AI:}
In general, deep-learning based models are known to process incrementally sophisticated features through the various layers \cite{zeiler2014visualizing}. While the lower level layers are responsible for feature extraction, the higher level layers are responsible for extraction of complex features, semantics and various other abstract concepts. Recently, very deep layered models (such as networks of large language models (LLMs) \cite{zhang2023wider}) have demonstrated  the capability of complex context-aware reasoning as well as alignment with human values. This is very similar to the \textit{why-vs-how} framework, where higher order layers can effectively perform reasoning as well as complex conceptualization. While certain aspects of abstractness have been individually assessed in these models (such as neutrality \cite{foroutan2022discovering} in language models), there is lack of studies that compares the ability of models to demonstrate the various aspects of abstractness parallelly, making it difficult to understand the true reasoning capabilities of AI models.

\subsubsection{\textbf{Creativity as a result of neural activation}}

The brain is a highly creative engine, where new ideas are triggered as a result of forming connections between fragments of information from memory \cite{herrmann1991creative}. This is primarily carried out in an independent internal network in the brain known as the \textit{default mode network}, which is primarily driven by the objective of optimizing future event prediction \cite{beaty2014creativity}.
Apart from this, an interesting characteristic of the creativity mechanism is explained by the \textit{spreading activation theory} \cite{collins1975spreading}. According to this theory, processing a particular concept in a neural node activates neighboring neural pathways, thus leading to creation of novel ideas \cite{schubert2021creativity}. 


\underline{Parallel with AI:}
From a computational perspective, creativity can majorly be divided into three categories \cite{franceschelli2021creativity}, each of which relates to different methods in AI. This includes 

\begin{itemize}

    \item \textit{Combinatorial creativity}, which relates to combining priors with sensory information in order to predict future events. This is seen in various probabilistic time modeling techniques \cite{neumann2019future}.    
    \item \textit{Exploratory creativity}, which relates to predicting totally new events through a trial-and-error approach. This can be seen in techniques such as reinforcement learning \cite{cohen2007reinforcement}, which optimize future event prediction by method of exploration. 
    \item \textit{Transformational creativity}, which converts sensory information to a new form by combining it with other information cues. This can be seen in various generative methods such as diffusion models \cite{croitoru2023diffusion} and adversarial networks \cite{li2022comprehensive}. 
\end{itemize}

These various methods explore different aspects of the creative process of the brain. For example, while combinatorial and exploratory methods in AI can be designed for effective future event prediction, they cannot process related information to generate novel ideas. On the other hand, transformational creativity can effectively emulate the \textit{spreading activation theory}, but has not been yet explored well enough for future event prediction. A more holistic approach towards creativity is hence required in this domain of research.

 \section{\textbf{Top-down processing}}
 Higher-order cognitive functions, such as anticipation, expectation, attention and memory play important roles in guiding, shaping and filtering the signals during perception. A sensory signal can behave differently in different settings as a result of top-down processing. A summary is provided in Figure \ref{fig:topics_covered} and Table \ref{tab: review}.

\subsection{\textbf{In neuroscience}}

\subsubsection{\textbf{Neuronal Synchronization as a guiding mechanism}}

Top-down processing refers to the influence of various factors such as context (sensory information related to a particular perceptual signal), and higher-order cognitive processes (such as attention and memory), on the outcome of perception, which guides thought in a particular direction \cite{rousay2023top}. Several biological theories indicate mechanisms relating to top-down processing in the brain. Some important theories include the \textit{neuronal synchronization theory} \cite{guevara2017neural}, suggesting that higher cognitive functions modulate the neural oscillations in different regions of the brain, which guides the neural pathway based on particular frequency channels matched by oscillations, leading to different outcomes in different situations. 

\underline{Parallel with AI:}
Theories of top-down processing from neuroscience, such as \textit{neuronal synchronization}, have not been well explored in AI. This theory associates different perceptual signals to a characteristic firing frequency, which helps in many cognitive properties such as integration of related features and coupling of different modalities. The current approach in AI is to treat each activation from the model as a discrete value, and all neuron activations are simultaneously generated with each forward pass. This leads to the treatment of AI models as one big black box. More research is required in this domain.

\subsection{\textbf{In psychology}}

\subsubsection{\textbf{Templates for guiding bottom-up processing}} 

All bottom-up processes are influenced by guiding cues from higher-level cognitive functions. The nature of these guiding functions is described by the \textit{schema theory} \cite{mcvee2005schema}, which proposes the existence of cognitive guides (schemas). Schemas are imaginary frameworks that encapsulate generalized ideas or object representations and are used as templates to guide perceptual processing. Schemas make processing efficient by providing similar sensory input signals with a predetermined perceptual path consisting of only relevant processes.

\underline{Parallel with AI:}
In the context of AI, the schema theory is related to concept representation in the deeper layers of the model. For example, the representation of similar objects and concepts should be closer to each other, irrespective of variations such as viewpoint or illumination. Similarly, it may be desirable to increase the distance between two related objects in a discriminatory setting. Understanding the feature representation space is hence necessary to properly model constraints as well as optimization goals in AI \cite{tamir2023machine}.  In \cite{chang2021concept},   a crude framework for optimizing deep learning models is presented,  based on concept representation analysis. 

However, there is still a lot of work that can be done in this field of research. Feature representation of complex, compositional, or abstract concepts is still not understood well. There is growing interest in this domain, especially in the field of explainable AI \cite{sokol2020interpretable, borrego2022knowledge}. 


\subsection{\textbf{In linguistics}}

\subsubsection{\textbf{Filtering competing activations}}



The association and simultaneous triggering of multiple related subconcepts involves selection among many competing alternatives in order to carry out effective semantic understanding. According to the \textit{competing alternatives theory} \cite{moss2005selecting}, competing concepts such as polysemous (words that have multiple definitions) and homonymous (words that sound the same but have different meanings) words are sequentially filtered through higher-order cues via a top-down feedback path. There are two major theories explaining selection between competing alternatives in the brain. The \textit{garden path theory} \cite{frazier1982making} proposes that sentences are processed incrementally, and the cognitively simplest definition is chosen as the interpretation, until it creates a conflict with the narrative of the discourse. Once, a conflict is encountered, an alternative interpretation is selected. The \textit{constrain based theory} 
 \cite{thompson1999effects}  alternatively proposes that this selection process is majorly statistical in nature, where the pathway with the highest activation is selected and processed further. This phenomenon is more or less consistent across all languages, as is observed through the study of implicit causality  \cite{hartshorne2013implicit} (i.e., the understanding of the source of an action without an explicit mention in a sentence). 

\underline{Parallel with AI:}
While the \textit{garden path theory} conserves working memory resources, the \textit{constrain based theory} is more resource-intensive, but highly efficient. This presents a efficiency-performance trade-off from a computational perspective. Depending on resource availability and performance requirements, transformer based language model families (such as GPT and BERT) can be modeled to exhibit both garden path based prediction\cite{jurayj2022garden, irwin2023bert}  as well as constrain based prediction \cite{garbacea2022constrained} in different task settings.
Further, the selection among competing concepts (across languages, tasks, etc.) can be selectively designed through targeted fine-tuning \cite{davis2021uncovering}. 
 
 \section{\textbf{Interpretation}} \label{interpretation}
 Effective decision making in the brain is a result of combining processed sensory information with other forms of knowledge, including prior knowledge, previous states of sensory signals as well as various kinds of heuristics. The overall interpretation mechanism of the brain can be visualized through Figure \ref{fig:topics_covered} and Table \ref{tab: review}.

\subsection{\textbf{In neuroscience}}

\subsubsection{\textbf{Predictive processes}}

According to a  prominent theory in neuroscience in the brain is the \textit{predictive coding theory} \cite{shipp2016neural, caucheteux2023evidence},  the brain continuously predicts future sensory signals in higher-level cortical areas, while also constantly optimizing prediction in new environments.  This mechanism allows for efficient perceptual processing by reducing the amount of sensory data required to make informed decisions. Predictive coding theory also acts as the primary method of learning representations of the world. 

\underline{Parallel with AI:} Current AI algorithms derive many ideas from the predictive coding mechanism, which postulates the general principles of prediction and learning in the brain \cite{kok2015predictive}. This includes hierarchical propagation of error signal \cite{mikulasch2023error}, similar to layerwise error propagation in backpropagation \cite{lillicrap2020backpropagation}, widely used in neural network optimization. Certain representations of the prediction error \cite{den2012prediction}, such as dopamine release, are also translated to objective functions in reinforcement learning \cite{richards2019deep}. Other forms of predictive error modeling, such as spiking error, can be widely seen in the body of  spiking neural network \cite{ghosh2009spiking}.  However, backpropagation operates by minimizing a global loss function that drives the entire model architecture. Predictive coding, on the other hand, provides means for learning based on available local neural activity alone \cite{rosenbaum2022relationship}. However, there is a body of work that has focused on enhancing the current backpropagation algorithm through predictive coding principles \cite{millidge2022predictive}.

Predictive coding mechanism in the brain is also characterized by recursive and bidirectional connections. Many classes  of AI models have properties of recursive connections (as in  recurrent neural networks \cite{salehinejad2017recent}) and bidirectionalism (as in bidirectional LSTMs \cite{graves2005bidirectional}). However, recurrent connections in AI models have majorly been implemented locally about a single layer. There are many scales of recurrent loops in the brain \cite{kawaguchi2017pyramidal}. These involve not only recurrent connections between layers, but between modules and also local node groups with individual layers. This needs to better explored in AI models. 

\subsubsection{\textbf{Optimization principles of interpretation}}

Predictive coding falls under a more general theory known as the \textit{free energy principle} \cite{friston2010free}, according to which, predictive optimization mainly focuses on reducing two types of energy costs of processing information, (a) \textit{variational free energy} \cite{friston2007variational}, related to processing unfamiliar sensory signals, and (b) \textit{expected energy}, related to the general prediction of future events \cite{millidge2021whence}.   


\underline{Parallel with AI:}
According to the free energy principle,  the brain tries to minimize \textit{variational free energy}, which aims to minimize the surprise when the brain comes across an unfamiliar input. e is executed by combining sensory input and prior information. Through this mechanism, the brain performs well on tasks such as few shot learning, and out-of-distribution detection, which are particularly challenging tasks for AI models \cite{lee2022brain}.  Many energy optimization problems have been formulated in AI to tackle the problem of out of distribution detection \cite{yang2021generalized}, as well as few shot learning \cite{lu2020learning}. The brain also tries to minimize \textit{expected free energy}, which aims to predict future events with maximal certainty, through bayesian prediction \cite{millidge2021whence}. This form of energy optimization is also frequently observed in the field of reinforcement learning \cite{ogishima2020combining}. There has also been work on frameworks to integrate both \textit{variational free energy} and \textit{expected free energy} in the context of bayesian modeling for AI model learning \cite{mazzaglia2022free}.

\subsection{\textbf{In psychology}}

\subsubsection{\textbf{Nature of prediction}}

Prediction in the brain is related to combining sensory input with contextual information derived from mental models or frameworks \cite{johnson2001mental} that are used to predict future outcomes. These mental models are continuously learned and manipulated based on the outcomes of perceptual events and other factors such as rewards.

A contemporary approach towards prediction in the brain is proposed by the \textit{dual theory process} \cite{gawronski2013dual}, which posits that the perceptual system consists of two parallel modes of thinking - a heuristical method (used to make fast inferences through cognitive shortcuts), and an active method of thinking( involving step-by-step reasoning and careful deliberation of various possible outcomes). The mix between the two types of responses is affected by factors \cite{kahneman2009conditions} such as time pressure (high-pressure tasks are carried out without much deliberation), expected rewards (lower reward tasks are likely to be handled with less deliberation), and experience (repetitive tasks are more likely heuristically performed).

\underline{Parallel with AI:}
 The dual theory process, which is well studied in cognitive sciences proposes an interesting framework for AI systems. Current AI models treat all input signal uniformly, through a systematic processing mechanism. However, there are some classical approaches that can be used to augment the current systematic reasoning methods with heuristical approaches, such as the Order of Magnitude Reasoning framework \cite{raiman1990order,mavrovouniotis1987reasoning,yip1996model}, and various similar reasoning frameworks \cite{dubois1990order, de1984qualitative}. The interplay between both types of modules may help AI systems to modulate between different time or resource constrained contexts. 

\subsubsection{\textbf{Inference}}

Along with various sensory clues, efficient decision-making also requires prior information gained through experience \cite{friston2003learning}. Inference in the brain is highly probabilistic in nature, where experiences that are more recent, and/or repetitive (thus represented by a larger neuronal population) have a higher effect on present perception \cite{pitkow2017inference}. While it is largely believed that the brain follows bayesian-like probabilistic inference \cite{shams2022bayesian}, recent theories also claim the possibility of other probabilistic models of the inference engine in the brain \cite{aitchison2017or}. 

\underline{Parallel with AI:}
Bayesian approaches have consistently shown improvement in many AI applications \cite{wang2020survey, wilson2020case}. However, the underlying information provided by prior (or underlying pieces of information aiding decision making, which is otherwise hard to learn with limited data) are not well understood, and heavily underutilized in the context of AI \cite{lu2022all}. Understanding the influence of data that informs further decisions in a hierarchical manner \cite{koh2017understanding} can allow for better reasoning in AI systems, which currently lack complex reasoning capabilities \cite{huang2022towards}. Better prior utilization can also improve learning efficiency, by reducing the dependency of current AI systems on large amounts of data for training \cite{muggleton2023hypothesizing}.

\subsection{\textbf{In linguistics}}

\subsubsection{\textbf{Semantic understanding}}
The definitions of various sentences have to be combined to derive semantic information for deep contextual understanding. According to the \textit{discourse representation theory} \cite{kamp2011discourse}, the brain integrates knowledge across multiple sentences by keeping track of the states of key entities (or ``centers" \cite{walker1998centering}). These states are referred to as situation states \cite{zwaan1998situation}. 
The \textit{event-indexing model} further divides situation states are into five major classes of entities, namely protagonist (the center theme of the discourse), temporality, causality, spatiality and intentionality \cite{zwaan1995construction}. During a sequence of events, one or more entities of the sitation state may be updated in the memory \cite{reynolds2007computational}. Related events (i.e., events that differ by a minimal number of entity classes) are more correlated in the memory than unrelated events.

\underline{Parallel with AI:}
Many language models \cite{tenney2019bert, clark2019does} demonstrate the ability to perform basic entity tracking (such as capturing coreference relations \cite{hobbs1979coherence}) through targeted training. While many datasets are specifically designed to train and evaluate models on entity tracking tasks \cite{chen2018preco, bamman2019annotated, uryupina2020annotating}, entity tracking is also an emergent property observed in large language models, as a result of pretraining on large text corpora. However, text-based training alone is incapable of training models for complex entity tracking. Complex entity tracking is observed as a by-product of training on code-based data (i.e. computer programs) along with text-based data \cite{kim2023entity}. This may be a result of the involvement of  structured and systematic transformation of multiple variables states in code. However, code cannot capture many non-trivial, unstructured linguistic semantics, which is why there is a need for text-based datasets that can train models on complex and simultaneous tracking of multiple entities in a discourse. Further, no dataset yet published classifies entities based on the \textit{event-indexing model}. 

\section{\textbf{Conclusion}} \label{conclusion}
 

Cognitive sciences clearly has many interesting insights to offer to perception in AI. In Table \ref{tab: review}, many potential existing techniques in AI have been highlighted that can bring about cognitive influence in perception systems in AI. These methods, however, are only a starting point for AI researchers, since many cognitively inspired methods currently existing in AI are tailored to specific settings and assumptions. Thus, they may not be generalizable to other problems where AI can be applied. Conversely, cognitive scientists also can immensely  benefit from adopting these methods to create more viable and potent cognitive architectures that can provide quantifiably more efficient intelligent systems.

\subsection{\textbf{Major gaps in cognitive AI research}}
Through the process of reviewing the state of cognitive AI, we came across a large number of themes from neuroscience, psychology and linguistics, that have no, or minimal counterparts in the field of AI. Here, we summarize, domain by domain, the major gaps in research, providing many potential research directions for AI researchers. 

\paragraph{\textbf{Research gaps in neuroscience-based AI}}
We observe two major themes in neuroscience, that are more or less unexplored in AI. 
\begin{itemize}
    \item Firstly, \textit{neural architecture and design}, which relates to the study of the wiring of the brain structurally provides some interesting insights. The brain is highly modular at various levels of hierarchy. Although modularity is recognized by the AI community, the application of modularity in mainstream research is limited to problem-specific architecture design or to separate different chains of processes \cite{amer2019review}.  We also observe that concepts like temporal coincidence, which points towards the ability to dynamically group similar signals without explicit architectural design is completely unexplored in AI. 

    \item Secondly, AI lacks insights in many \textit{neural functions}, which study various functions that arise specifically out of neuronal cell grouping. Object Permanence, while one of the most fundamental functions of cognitive perception, is crudely addressed in the field of AI, primarily through problem-specific data-driven approaches. The neural nature of object permanence in the human eye highlights that this function is less learned, and more mechanistic in nature. Similarly, tonotopicity, which indicates towards the ability to filter specific kinds of signals is not well analysed in current models in AI. 
\end{itemize}

\paragraph{\textbf{Research gaps in psychology-based AI}}
Similar to neuroscience, we also observe two major themes in psychology, that have limited reach in AI research. 
\begin{itemize}

    \item AI can borrow many interesting insights from the cognitive mechanisms underlying \textit{contextual processing}, or the function that augments the knowledge derived from sensory signals by incorporating prior knowledge cues. AI still lacks efficient methods to utilize goal-oriented feedback connections, which also extends to the ability to anticipate future signals. Signal denoising is, again, highly data-centric in mainstream AI, unlike the computationally inexpensive mechanistic mechanism of the brain. On the contrary, properties such as signal selectivity (or the property of discarding irrelevant signals) is a task to be solved by specific training schemas. 

    \item Psychology also has many insights to offer into \textit{architectural design for AI models}, such as dynamic sharing and integration of cognitive resources. As such, progressive modularization  and integration of modalities is not well explored in the field of AI. 
\end{itemize}

\paragraph{\textbf{Research gaps in linguistics-based AI}}
Finally, we also observe two major themes in linguistics, that are underexplored in AI, despite the huge interest in the field due to the rise of generative language models.
\begin{itemize}
    \item Many known limitations in natural language processing \cite{khurana2023natural} can be solved by encoding better \textit{linguistic structures} in models. A dynamic vocabulary structure, for example, is a challenge to implement in current AI models, as changing representations of the vocabulary requires one to retrain the entire model. However, the brain is capable of enhancing the internal vocabulary representations without affecting, or specifically downgrading, the quality of subsequent sensory representations in the linguistic pathway of perception. Similarly, AI still has not explored the benefits of a structure that emulates the phonological buffer in the brain. 
    
    \item \textit{Language generation semantics} is still an open problem in AI, with lots of  interest in the research community. In general, the most sophisticated language systems still suffer from the lack of creativity \cite{franceschelli2023creativity}. The reason for the shortcoming, which relates to the underlying design of autoregressive models currently used, directly extends to many other problems including hallucinations (or giving false information)  \cite{ye2023cognitive} and long-range coherence \cite{naveed2023comprehensive}. Studying the neural activation method in the brain behind the creative mechanism of the brain is a worthwhile research direction for researchers in AI. 
\end{itemize}

While each of the above-mentioned topics have a huge potential upside in research, an overarching characteristic that all research in cognitive AI (which requires the collaborative effort of both AI scientists and cognitive scientists from all related sub-domains) should avoid is designing methods and modeling techniques tailored for specific problems. Such an approach makes intelligent systems prone to pitfalls that blocks the path of AI towards general intelligence. In this spirit, a few directions of research should be explored actively alongside cognitive AI systems that aim to increase model efficiency and performance. 

\subsection{\textbf{Path towards general intelligence: future directions of research}}

\begin{figure}
    \centering
    \includegraphics[width=\linewidth]{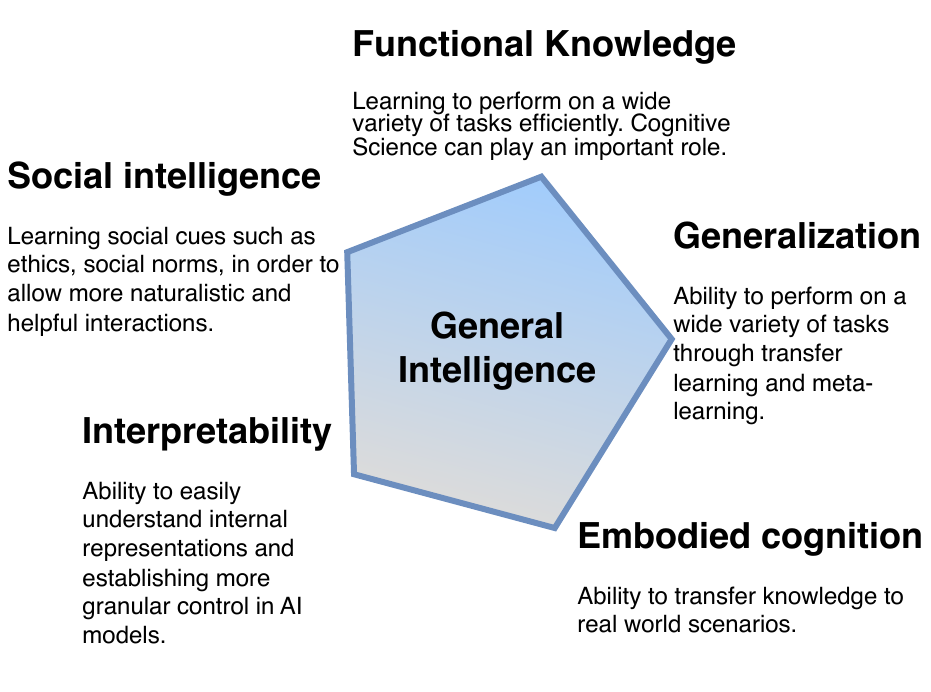}
    \caption{Schema for General Intelligence}
    \label{fig:AGI schema}
\end{figure}

The pursuit of general intelligence \cite{goertzel2014artificial} has been the ultimate motivation of the field of AI since its inception. Cognitive AI is, however, only one facet of general intelligence. The purpose of studying cognitive principles and borrowing inspiration from them into AI is to improve the learning capacity and efficiency (in energy, performance, comprehension of the environment and making better logical decisions). A truly intelligent system has many other aspects that must be given equal importance in the grand set of motivations for AI research. Overall, general intelligence is constituted of five major elements -- \textbf{functional knowledge} (that may be learned through cognitive sciences), \textbf{ability to generalize} to a broad spectrum of tasks and data, \textbf{embodied cognition}, \textbf{interpretability} and \textbf{social intelligence}. These aspects of general intelligence can be visualized in Figure \ref{fig:AGI schema}. As such, research efforts in AI as well as cognitive sciences should focus on these parallelly. While we discussed cognitive AI efforts in depth, here we describe in brief some future directions of research for holistically developing general intelligence through advanced perception.

\subsubsection{\textbf{Generalization}}
One of the hardest challenges in AI is to generalize performance of AI models across various domains without having to explicitly train models many tasks \cite{al2020generalizing}. Generalization is typically achieved to limited extents through the technique of \textit{transfer learning}\cite{zhuang2020comprehensive}. Many problems remain unsolved in the domain of transfer learning, including domain gap (where the target domain is not benefited by the knowledge of the source domain), and negative transfer \cite{zhang2022surveynegative}, which hinders retrieval of knowledge from the target domain. In order to better generalize to real world knowledge, models need to be regularized using psychophysical data \cite{duan2022survey}, that helps models achieve human-like perception and transfer knowledge in a practically useful manner.

\subsubsection{\textbf{Embodied cognition}}
Embodiment involves teaching AI agents to interact meaningfully with their surroundings with real-world constraints such as physics and, rather than a simulated interactions \cite{chrisley2003embodied}. Embodied AI is still in very primitive stages, where models can only handle narrow tasks such as visual object navigation \cite{duan2022survey}, and fail when extended beyond simple constraints. Cognitive science can especially aid embodied AI enhance perception beyond simple tasks by defining a structure for embodied principles on various levels of human interaction with the environment \cite{coello2015embodied}. 

\subsubsection{\textbf{Interpretability}}

In general, the learning patterns and mechanism of deep learning models is not well understood, which leads to the black-box nature of the large parametric models of today \cite{castelvecchi2016can}. Understanding the internal representations of AI models is crucial to gain control over the learning mechanisms of models and their behavior in different settings. This particular problem is explored in depth in the domain of explainable AI \cite{xu2019explainable}. The representation of sensory signals in perception can be better understood by taking into account cognitive schemas that separate the various components of explainability explicitly in an intuitive manner \cite{zhang2022towards}.

Apart from understanding parametric representations, having control over model training is equally important in order to guide learning of models in the right direction. This domain, commonly known as meta-learning \cite{vanschoren2019meta} is an active area of research in AI. For general AI, a unified understanding of meta-learning is required, and should be studied in depth going forward.

\subsubsection{\textbf{Social intelligence}}
An indispensable factor in human-like cognition is incorporating social dynamics in AI models, including understanding of ethics, social norms, social equality, naturalistic communication, and evolution of semantic usage and meaning over time \cite{van2021sustainable}. As AI models become more accessible and powerful, the potential to bring harm to society increases as well \cite{gratch2022power}. This problem cannot be simply solved by the power of data scaling, and needs to be addressed through putting down foundations of  societal structures carefully in AI models. These foundations need to emerge at various levels, including at global, national and communal level \cite{tomavsev2020ai}.

\bibliographystyle{unsrt}
\bibliography{main}
\end{document}